\documentclass[letterpaper, 10 pt, journal, twoside]{ieeetran}

\usepackage{cite}
\usepackage{epsfig}
\usepackage{siunitx}
\usepackage{hyperref}
\usepackage{comment} 
\usepackage{graphics}
\usepackage{amsmath}
\usepackage{bm}
\usepackage{colortbl}
\usepackage{mathtools}
\usepackage{amsmath,amssymb}
\usepackage{soul}
\usepackage{xcolor}
\usepackage{comment}
\usepackage{threeparttable}
\usepackage{booktabs}

\DeclareRobustCommand{\uvec}[1]{{%
		\ifcsname uvec#1\endcsname
		\csname uvec#1\endcsname
		\else
		\bm{\mathbf{#1}}%
		\fi
}}
\renewcommand{\baselinestretch}{.96} 
\begin{document}
\markboth{IEEE Robotics and Automation Letters. Preprint Version. Accepted March, 2025}
{Kanada \MakeLowercase{\textit{et al.}: Joint-repositionable Inner-wireless Planar Snake Robot}: } 

\author{Ayato Kanada$^{1*}$ Ryo Takahashi$^{2*}$ Keito Hayashi$^{1}$ Ryusuke Hosaka$^{1}$  Wakako Yukita$^{2}$ Yasutaka Nakashima$^{1}$ Tomoyuki Yokota$^{2}$ Takao Someya$^{2}$ Mitsuhiro Kamezaki$^{2}$ Yoshihiro Kawahara$^{2}$ Motoji Yamamoto$^{1}$%
\thanks{Manuscript received: November 25, 2024; Revised:
February 5, 2025; Accepted: March 2, 2025.}
\thanks{This paper was recommended for publication by
Editor Yong-Lae Park upon evaluation of the Associate Editor and Reviewers’ comments. This work was mainly supported by JST JPMJAX21K9, JPMJAX23K6, and JSPS 22K21343. \it(*Corresponding and co-first authors: Ayato Kanada and Ryo Takahashi)}
\thanks{
$^1$A. Kanada, K. Hayashi, R. Hosaka, Y. Nakashima, M. Yamamoto are with the Department of Mechanical Engineering, Kyushu University, Fukuoka, Japan, 819-0395. 
{\tt\small kanada, hosaka, hayashi, nakashima, yama@ce.mech.kyushu-u.ac.jp}}%
\thanks{
$^{2}$R. Takahashi, W. Yukita, T. Yokota, M. Kamezaki, Y. Kawahara, are with Graduate School of Engineering, The University of Tokyo, Tokyo, Japan, 113-8656. 
{\tt\small takahashi, kamezaki, kawahara@akg.u-tokyo.ac.jp, yukita, yokota@ntech.t.u-tokyo.ac.jp, someya@ee.t.u-tokyo.ac.jp}} %
\thanks{Digital Object Identifier (DOI): see top of this page.}
}

\title{{\fontsize{23.4pt}{0pt}\selectfont Joint-repositionable Inner-wireless Planar Snake Robot}} 

\maketitle

\begin{abstract}
Bio-inspired multi-joint snake robots offer the advantages of terrain adaptability due to their limbless structure and high flexibility.
However, a series of dozens of motor units in typical multiple-joint snake robots results in a heavy body structure and hundreds of watts of high power consumption.
This paper presents a joint-repositionable, inner-wireless snake robot that enables multi-joint-like locomotion using a low-powered underactuated mechanism.  
The snake robot, consisting of a series of flexible passive links, can dynamically change its joint coupling configuration by repositioning motor-driven joint units along rack gears inside the robot.
Additionally, a soft robot skin wirelessly powers the internal joint units, avoiding the risk of wire tangling and disconnection caused by the movable joint units. 
The combination of the joint-repositionable mechanism and the wireless-charging-enabled soft skin achieves a high degree of bending, along with a lightweight structure of 1.3 kg and energy-efficient wireless power transmission of 7.6 watts.
\end{abstract}

\section{INTRODUCTION}

\begin{figure}[t!]
	\begin{center}
		\includegraphics[width=0.98\columnwidth]{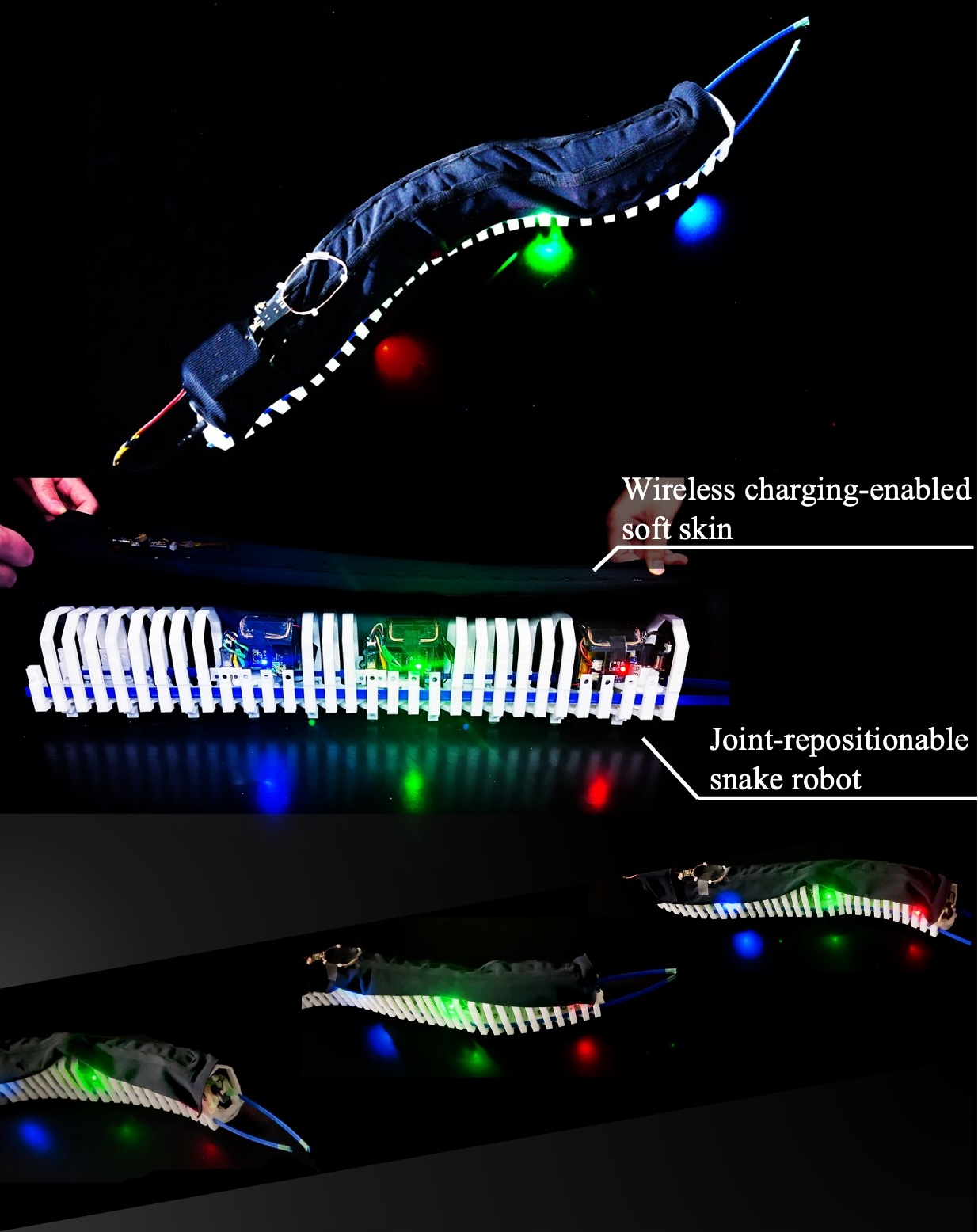}
	\end{center}
	\caption{\textbf{Concept of joint-repositionable inner-wireless planar snake robot}. Our robot enables multi-joint-like locomotion while remaining low-powered and lightweight structure. Inside the robot, joint-repositionable units can move freely, enabling to construct the various joint coupling. The units are powered wirelessly through a soft robot skin.}
	\label{fig:overview}
\end{figure}

Biological snakes obtain superb locomotion capabilities in rough terrains by adaptively bending their flexible bodies with a high degree of freedom~\cite{hoffstetter1969biology}. 
Multi-joint snake robots inspired by snakes in nature have demonstrated wider terrain adaptability in constrained, unstructured environments over wheeled or legged robots~\cite{pettersen2017snake}.
Despite their adaptive robotic locomotion, efficient and continuous locomotion with the snake robot remains a challenge.
The multi-joint mechanical design based on a series of dozens of motor-driven joints results in a heavy structure and hundreds of watts of power consumption due to the challenges associated with motor miniaturization~\cite{liu2021review}.
In the development of lightweight, energy-efficient snake robots, previous work has focused mainly on underactuated joint systems such as wire-driven joint bending or pneumatic soft actuation of the joints~\cite{qi2020novel,branyan2017soft,wan2023design}.
However, the design constraints of wires and pneumatic chambers restrict the variety of robot postures.

\begin{figure*}[t!]
	\begin{center}
		\includegraphics[width=0.98\textwidth]{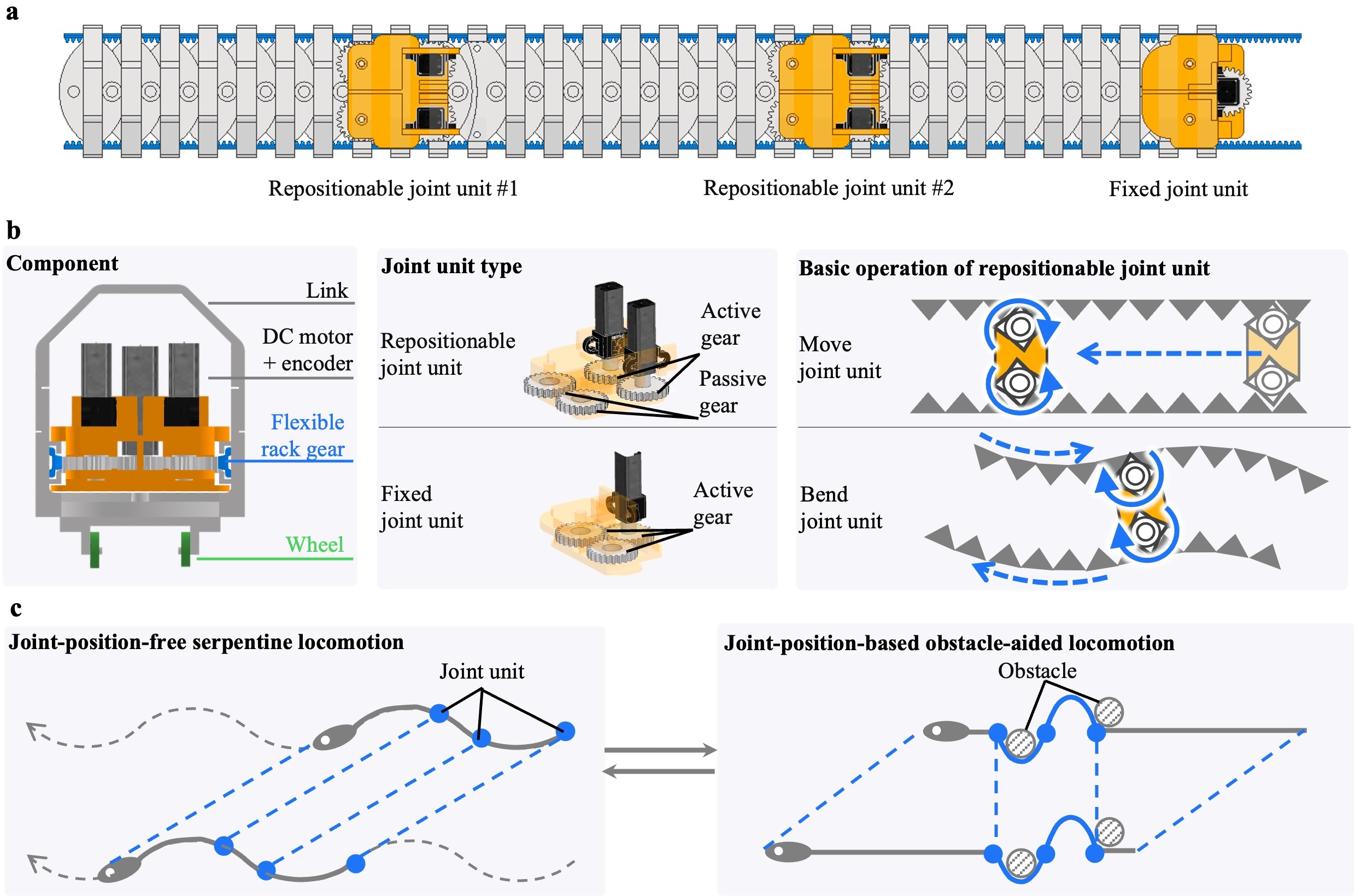}
	\end{center}
	\caption{\textbf{Design and operation of a joint-repositionable, inner-wireless planar snake robot.} (a) Schematic of the joint-repositionable snake robot, including repositionable joint units and a fixed joint unit. (b) Detailed view of the joint unit components, including a DC motor with an encoder, a flexible rack gear, and a wheel. The joint unit types are categorized as repositionable and fixed, with the basic operations of moving and bending depicted. (c) Two types of locomotion strategies for adapting to different environments: joint-position-free serpentine locomotion for obstacle-free, narrow terrains, and joint-position-based, obstacle-aided locomotion.}
	\label{fig:joint-design}
\end{figure*}

This article presents a new class of underactuated snake robots ---\textit{joint-repositionable inner-wireless snake robot}---.
Unlike the typical fixed joint coupling structure composed of dozens of motor-based joint units and rigid links, we introduce a joint-repositionable design, dynamically changing the joint coupling configuration by moving at least two motor-driven joint units inside the robot.
Our mathematical model and physical experiments demonstrate that the joint-repositionable mechanism realizes two types of untethered snake-like locomotion with up to \SI{3.6}{\W} power.
Additionally, a wireless charging-enabled soft robot skin can prevent the internal power cable from tangling with the moving joint units.
The optimization of wireless charging experimentally demonstrates watt-class wireless power transmission with over $50\%$ efficiency while avoiding electromagnetic interference with the metallic motors in the joints.
Overall, the joint-repositionable robot, wirelessly powered by the robot skin, enables basic multi-joint-like snake movements while maintaining a lightweight design of \SI{1.3}{\kg}~(motors: \SI{62.5}{\g}, battery: \SI{100}{\g}) and energy-efficient use through \SI{7.6}{\W} wireless charging~(see \autoref{fig:overview}).

\section{Robot Design and Implementation}

\subsection{Joint-repositionable Mechanism}

\begin{figure*}[ht!]
	\begin{center}
        \includegraphics[width=0.98\textwidth]{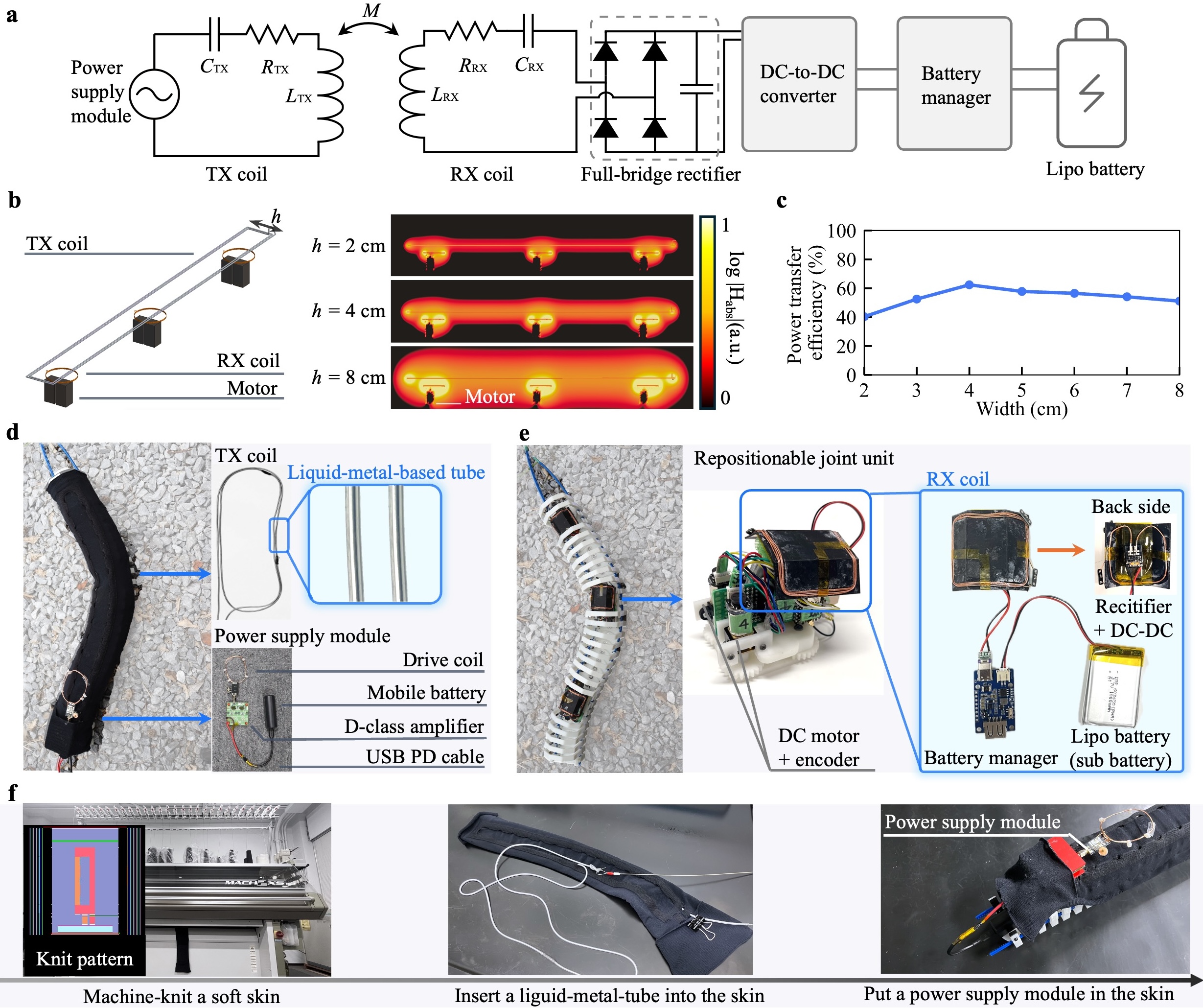}
	\end{center}
	\caption{\textbf{Design overview of a wireless-charging-enabled soft robot skin.} (a) Circuit diagram of the soft robot skin.(b) Simulated inductive field of the soft robot skin model, and (c) power transfer efficiency for the soft robot skin geometry. (d) Photograph of the soft robot skin composed of a liquid-metal-based transmitter coil. (e) Photograph of joint units connected to an RX coil. (f) Fabrication process of the soft robot skin.}
	\label{fig:skin-design}
\end{figure*}

The joint-repositionable mechanism consists of an exoskeleton body with serially connected links, passive two-way wheels, two flexible rack gears passing through the links, and $N$ motor-driven joint units that travel along the body (see \autoref{fig:joint-design}a for $N=3$). 
The two rack gears pass through grooves in each link, while their front ends are fixed to the first link, and their rear ends are left free. 
Among $N$ joint units, $N-1$ are joint-repositionable joint units, which consist of two motors that move along the robot body, and one is a fixed joint unit, which consists of one motor fastened at the robot's rear. 

The mechanical design and basic operation of the two types of joint units are illustrated in \autoref{fig:joint-design}b. 
The two motors in the repositionable joint unit mainly control the robot's locomotion. 
When the motors rotate in the same direction, one rack gear moves forward while the other moves backward, creating an S-shaped bend. 
In contrast, the rotation in opposite directions allows the movable unit to move back and forth along the rack gears. 
Two active gears connected to the motors is used to transfer the rotation force of the motor to the rack gears, while passive gears are used for hooking the movable units. 
The fixed joint unit for supporting the bending by the repositionable joint units, uses a single motor to synchronously rotate the two flexible racks.

Based on its basic operation, our robot provides two types of locomotion~(see \autoref{fig:joint-design}c). 
The first approach is a joint-position-free serpentine locomotion, smoothly navigating in obstacle-free narrow terrains. 
This locomotion eliminates the need for precise joint positioning, thereby reducing operational time. 
The second approach is a joint-position-based obstacle-aided locomotion, slowly crawling over surrounding obstacles in the terrain.
Adaptive joint positioning is necessary to fit the robot's shape to C-shaped obstacles.
The detail control strategy is described in \S~\ref{sec:model_and_control}.
In total, the joint-repositionable design enables snake-like locomotion while drastically reducing the number of necessary motors.

\subsection{Wireless Charging-enabled Robot Skin}

The repositionable joint units must be lightweight and freely movable. 
While equipping small batteries in the joint units allows for untethered and fast movement, the short lifespan becomes the issue. 
Conversely, the direct connection of power supply cables to the joint units causes wire tangling or breakage due to joint movement. 
Therefore, we employ a wireless charging approach, which transmits power from the external power module to the untethered joint units.
Specifically, coil-based inductive charging can efficiently transmit \si{\W}-class power, compared to other wireless charging approach like electromagnetic wave and capacitve coupling~\cite{WPT2024review}.

\begin{figure*}[ht!]
	\begin{center}
		\includegraphics[width=0.98\textwidth]{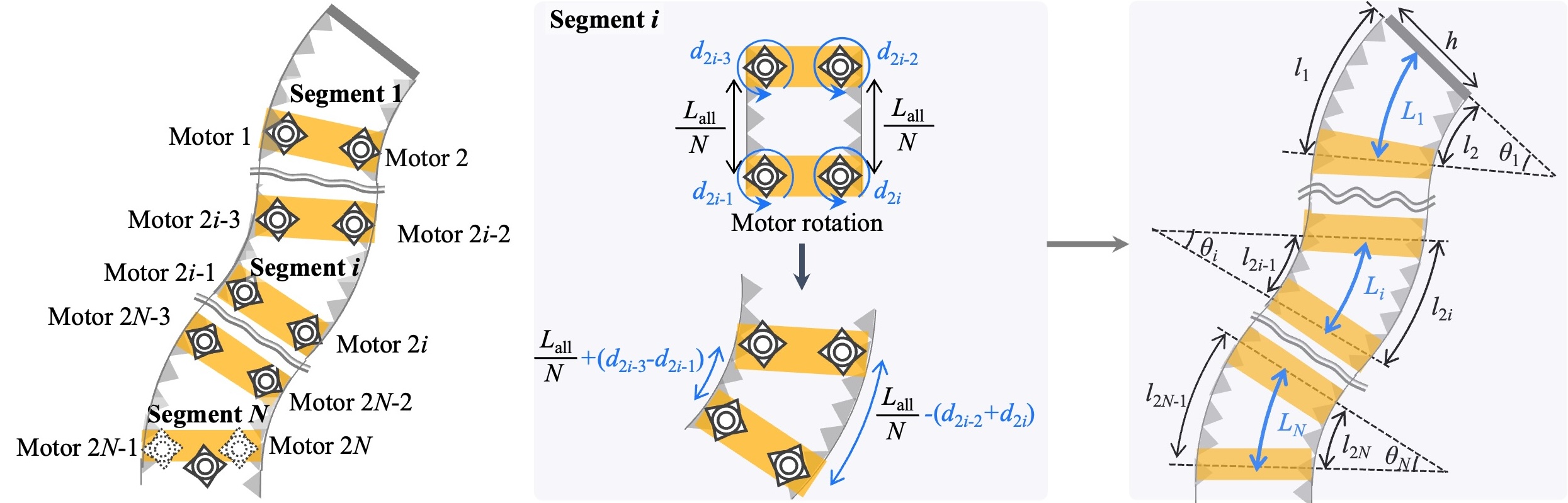}
	\end{center}
	\caption{\textbf{Schematic of the variable-length arc-shaped joint model.} Our robot, comprising $N$ joint units, is represented as a sequence of arcs with uniform curvature. The rotational movement of the motors ($d$) changes the length ($L$) and angle ($\theta$) of each arc segment.}
	\label{fig:model}
\end{figure*}

\autoref{fig:skin-design}a shows the circuit design of the wireless charging-enabled soft robot skin.
The skin consists of a transmitter (TX) helical coil connected to a \SI{6.78}{\MHz} power supply module and three receiver~(RX) spiral coils, each connected to one of the three joint units.
Through the inductive coupling, the TX coil wirelessly transmits AC power to the RX coil, which converts the received AC power into DC power via a full-bridge rectifier and a DC-DC converter, and then stores the energy in a battery.

Optimizing the TX helical coil is necessary because the metallic motors can potentially interact with the inductive field. 
The wireless power delivery distance from the helical coil is mainly determined by its shorter side length~(\textit{i.e.}, width). 
To understand the relationship between coil width and maximum power transfer efficiency~\cite{Zargham2012effcalc}, we conducted an electromagnetic simulator (FEKO, HyperWorks).
The simulation model includes i) a helical TX coil with a length of \SI{55}{\cm}, $2$ turns, and a conductivity of \SI{3.4d7}{S/m}, ii) a \qtyproduct{4x4}{\cm} $4$-turn RX coil, and iii) cube-shaped simplified stainless motor units. 
The distances between the TX/RX coils and the RX coil/motor are \SI{1}{\cm} and \SI{0.5}{\cm}, respectively, and the width of the TX coil varies from \SI{2}{\cm} to \SI{8}{\cm} in increments of \SI{1}{\cm}~(see \autoref{fig:skin-design}b).
The results indicate that a wider coil interacts with the motors, while a narrower coil fails to deliver a strong inductive field to the RX coil~(see \autoref{fig:skin-design}c). 
The \SI{4}{\cm} width is chosen for efficient power transmission.

\subsection{Prototype Implementation}

\autoref{fig:skin-design}de illustrates the prototype mainly consisting of the stretchable TX coil and the joint-repositionable mechanism connected to the RX coil, respectively.
The TX coil is fabricated by inserting the liquid-metal-based stretchable tube with the inner/outer diameter of \num{2}/\SI{3}{\mm} into a knitted textile, similar to \cite{takahashi2022meander}~(see \autoref{fig:skin-design}f).
Galinstan~(Ga68.5In21.5Sn10, Sichuan HPM) is selected as the liquid metal for its low melting point~(\SI{-19}{\degreeCelsius}), high electrical conductivity~(\num{3.5d7}\si{S/m}), and low toxicity.
For wireless power transmission, the TX coil is connected to the power supply module consisting of \SI{6.78}{\MHz} D-class amplifier~(EPC9065), \SI{22.5}{\W} DC mobile battery, and a USB PD cable with \SI{9}{\V} output. 
A $2$-turned drive coil is electrically connected to the amplifier, enabling efficient wireless power delivery from the amplifier to the TX coil.
By contrast, the \qtyproduct{4x4}{\cm} RX coil consists of a $4$-turn copper wire with the diameter of \SI{0.5}{\cm}, a full-bridge rectifier circuit with a DC-DC converter~(PMEG6010 and MIC29150-5.0), and a battery manager~(Lipo Rider Plus) connected to a sub small \SI{100}{mAh} Lipo battery.
The resonant frequency of the TX/RX coils was tuned at \SI{6.78}{\MHz} with the distributed chip capacitors~($C_{\rm TX}, C_{\rm RX}$)~\cite{ryo2024picoRing,sato2025mems}.
The inductance~($L_{\rm TX}, L_{\rm RX}$) and resistance~($R_{\rm TX}, R_{\rm RX}$) of the TX/RX coils at \SI{6.78}{\MHz} are \num{2.5}/\SI{2.1}{\uH}, \num{1.5}/\SI{0.7}{\ohm}, respectively.
Note that the joint units mostly rely on the wireless power from the mobile battery, but the sub battery serves as a backup.
The battery manager switches the two power supply channels.

The joint-repositionable mechanism with a total length of \SI{0.6}{\m} includes 3D-printed $30$ passive links with passive two-way wheels with a diameter of \SI{1.2}{\cm}, two flexible rack gears, and three joint units. 
The distance between the joints can be minimized up to \SI{20}{\mm}, enabling $30~(=\SI{0.6}{\m}/\SI{30}{\mm})$ joint number.
The joint unit has one or two \SI{0.5}{Nm} DC motors~(1000:1 Motor, Pololu) connected to a magnetic encoder~(3499, Pololu). 
The encoder measures the flexible rack length moved by the motor~($d$ in \autoref{fig:model}).
The total weight and maximum power consumption are \SI{1.3}{\kg}~(joint-repositionable unit: \SI{0.14}{\kg}, fixed joint unit: \SI{0.11}{\kg}, robot skin with a battery: \SI{0.3}{\kg}) and \SI{3.6}{\W}, respectively.
We experimentally confirmed the soft skin with \SI{7.6}{\W} DC input sends up to \SI{3.6}{\W} DC power to three joint units, respectively, during the robot's locomotion.

\autoref{tab:comparison1} shows a mechanical comparison of mobile snake robots from previous studies. Prior snake robots exhibit a trade-off between weight and joint length; robots with fewer motor-driven joints are lightweight, but the longer joint spacing resulting from the reduced number of joints restricts postural diversity, and vice versa. In contrast, the joint-repositionable snake robot with three joint units can be categorized among lightweight ($\approx$ 1 kg) groups, while also achieving a multi-joint structure with both short joint spacing (20 mm) and over 20 degrees of freedom. Among snake robots, our joint-repositionable robot, which is lightweight, low-powered, and highly articulated, has the potential for untethered, long-term autonomous operations in unstructured environments, including unpaved roads, rocky terrains, and collapsed buildings.

\begin{table}[h!]
    \centering
    \begin{threeparttable}
    \setlength{\tabcolsep}{4.1pt} 
    \renewcommand{\arraystretch}{1.1} 
    \caption{Mechanical comparison of mobile snake robots.}
    \label{tab:comparison1}
    \begin{tabular}{l|llll}
    \textbf{Snake robot} & \textbf{\begin{tabular}[c]{@{}l@{}}Length\\(m)\end{tabular}} & \textbf{\begin{tabular}[c]{@{}l@{}}Weight\\(kg)\end{tabular}} & \textbf{\begin{tabular}[c]{@{}l@{}}Joint spacing\\(mm)\end{tabular}} & \textbf{\begin{tabular}[c]{@{}l@{}}Joint number\\(DoF)\end{tabular}} \\
    \hline\hline
    AmphiBot I~\cite{AmphiBot} & $0.56$ & $\mathbf{0.7}$* & $70$ & $8$ \\
    ACM R5 & $1.7$ & $7$ & $189$ & $9$ \\
    AIKO~\cite{transeth2008snake} & $1.5$ & $7$ & $75$ & $\mathbf{20}$ \\
    Bayraktaroglu~\cite{Bayraktaroglu} & $0.65$ & $\mathbf{0.9}$ & $65$ & $10$ \\
    Unified Snake~\cite{unified} & $0.94$ & $\mathbf{2.9}$ & $59$ & $\mathbf{16}$ \\
    Takemori~\cite{Takemori} & $2.4$ & $15$ & $80$ & $\mathbf{30}$ \\
    PEA-Snake~\cite{PEA-snake} & $0.5$ & $1.7$ & $104$ & $5$ \\
    \rowcolor{gray!20}
    \textbf{Ours} & $0.6$ & $\mathbf{1.3}$ & $\mathbf{20}$ & $\mathbf{30}$ \\
    \bottomrule
    \end{tabular}

    \begin{tablenotes}[flushleft]
        \item \begin{minipage}[t]{\linewidth} 
        Note: *The weight of AmphiBot I is estimated referring to AmphiBot II. \textbf{The bold number} means the robot is either lightweight or highly articulated.
        \end{minipage}
    \end{tablenotes}
    \end{threeparttable}
    \label{tab:comparison2}
\end{table}

\section{Modeling and Control}
\label{sec:model_and_control}

\subsection{Variable-length Arc-shaped Joint Model}
\label{sec:model}

This section describes the generalized kinematic model of our snake robot referred to as the variable-length arc-shaped joint model.
Our joint-repositionable snake robot is modeled as a series of multiple arcs with variable lengths.
Here, we estimate both the position and angle of the repositionable joints based on the motor rotations in the joint units.
First, we assume the joint model with $N-1$ repositionable joint units and $1$ fixed joint unit can be divided into $N$ arc segments. 
Note that the first segment is separated by only the $1$-st joint unit.
Based on the piecewise constant curvature assumption, each segment is simplified as a circular arc with constant curvatures, enabling simple mathematical modeling of the robot's shape. 
Specifically, the $i$-th arc segment ($i=1, 2, ..., N$) between ($i-1$)-th and $i$-th joint units can be modeled by three lengths (centerline: $L_i$, left sideline: $l_{2i-1}$, right sideline: $l_{2i}$ and the arc angle $\theta_i$~(see \autoref{fig:model}).
The $i$-th repositionable joint unit has two ($2i-1$)-th and $2i$-th motors.
For simplicity, the $N$-th fixed joint unit is assumed to be composed of the ($2N-1$)-th and $2N$-th motors synchronously rotating in the reverse direction against the actual motor.
With this setup, $L_i$, $l_{2i-1}$, $l_{2i}$, and $\theta_i$ can be expressed as follows with a robot width $h$:
\begin{align}
   L_i &= \frac{l_{2i-1}+l_{2i}}{2} \notag,~~L_{\rm all} = \sum_{i=1}^N L_{i}=\text{Constant} \notag \\
   \theta_i &= \frac{l_{2i-1}-l_{2i}}{h}  \label{eq:definition}
\end{align}
where $L_{\rm all}$, the total length of the robot centerline, is constant.

\begin{figure*}[t!]
	\begin{center}
		\includegraphics[width=0.98\textwidth]{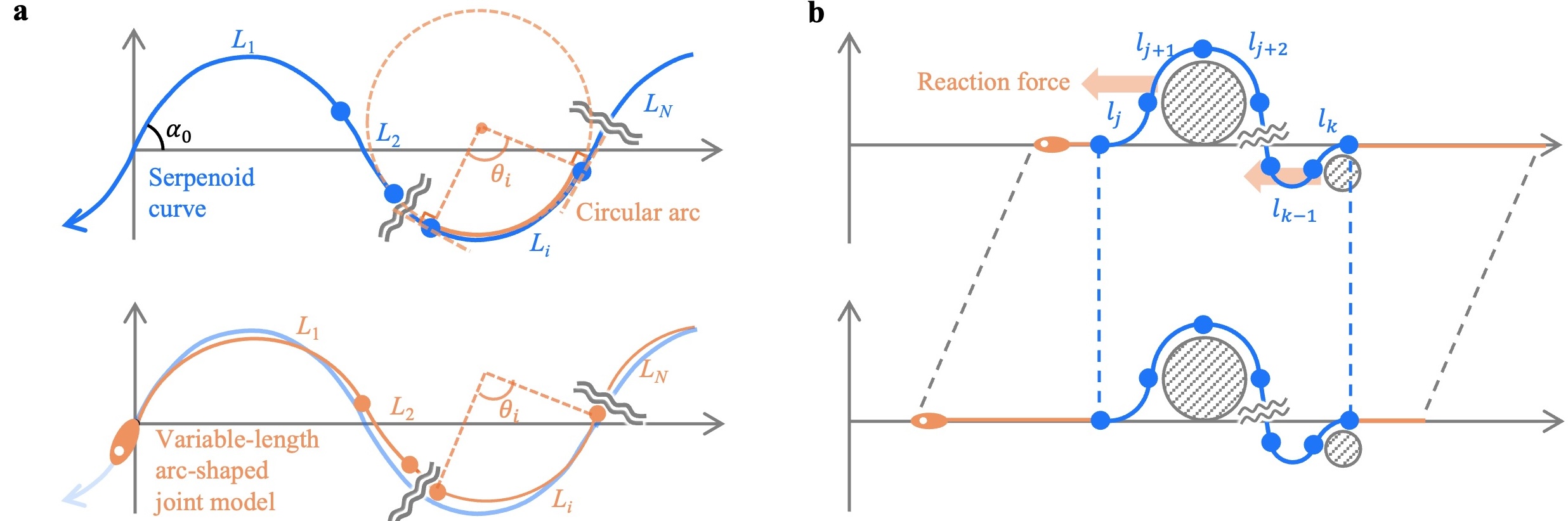}
	\end{center}
	\caption{\textbf{Locomotion control strategy using the variable-length arc-shaped joint model.} (a) Illustration of joint-position-free serpentine locomotion along a blue-colored serpenoid curve. (b) Illustration of joint-position-based obstacle-aided locomotion, highlighting orange-colored counter forces. When the fixed shape encounters an obstacle, the resulting counterforce propels the robot forward while keeping its blue-colored shape.}
	\label{fig:control}
\end{figure*}

Next, we model the change in $L_i$ caused by the motors in the $i$-th segment. 
We assume that the initial robot shape is straight and that the motor units are evenly positioned.
With this initial configuration, each rack gear length is assumed to be $L_{\rm all}/N$. 
The rotations of the ($2i-1$)-th and $2i$-th motors in the $i$-th segment increase $l_{2i-1}$ and $l_{2i}$ by a length of $d_{2i-1}$ and $d_{2i}$, respectively.
Since the  $2i-1$ and $2i$-th motors also change the length of the ($i+1$)-th segment, $l_{2i-1}$ and $l_{2i}$ can be expressed as follows in terms of $L_{\rm all}$, $N$, and $d$.
\begin{align}
    l_{2i-1} &= \cfrac{L_{\rm all}}{N} + (d_{2i-3} - d_{2i-1}) \notag\\
    l_{2i} &= \cfrac{L_{\rm all}}{N} - (d_{2i-2} - d_{2i}) \label{eq:d_to_l}
\end{align}
where $d_{-1} = d_{0}\coloneqq 0$ and $d_{2N-1} = d_{2N}$.
Note that the counterclockwise rotation of the motor is defined as a positive $d$.
Therefore, the conversion of $d_{2i-3}\cdots d_{2i}$ into $L_i$ and $\theta_i$ can be expressed as follows with \autoref{eq:definition} and \autoref{eq:d_to_l}:
\begin{align}
    L_i &= \cfrac{d_{2i-3} - d_{2i-2} - d_{2i-1}  + d_{2i}}{2} + \cfrac{L_{\rm all}}{N} \notag\\
    \theta_i &= \cfrac{d_{2i-3} + d_{2i-2} - d_{2i-1} - d_{2i}}{h}  \label{eq:d_to_Ltheta}
\end{align}
\autoref{eq:d_to_Ltheta} can be converted as follows: 
\begin{align}
    d_{2i-1} &= \sum_{j=1}^i \left(-L_j - \frac{\theta_j h}{2} + \cfrac{L_{\rm all}}{N}\right) \notag\\
    d_{2i} &= \sum_{j=1}^i \left(L_j - \frac{\theta_j h}{2} - \cfrac{L_{\rm all}}{N}\right)
    \label{eq:Ltheta_to_d}
\end{align}
\autoref{eq:Ltheta_to_d} shows the conversion of the robot configuration ($L$ and $\theta$) into motor actuation parameters ($d$).

\subsection{Joint-position-free Serpentine Locomotion}
\label{sec:Joint-position-free Serpentine Locomotion}

Among various terrestrial locomotion modes in snakes~\cite{liu2021review}, planar snake robots mainly use serpentine movement and obstacle-aided locomotion~(see \autoref{fig:control}). 
The serpentine movement enables snake robots to move smoothly in narrow terrains by bending their entire bodies along sinuous curves such as a serpenoid curve~\cite{Serpentine2002Sato}. 
By contrast, obstacle-aided locomotion allows the snake robots to push against terrain irregularities and narrow passage walls~\cite{transeth2008snake}.
Unlike well-established fixed-length arc-shaped joint models~\cite{774054,yamada2010approximations, Serpentine2002Sato}, no control method has yet been developed for a variable-length arc-shaped joint model.

\begin{figure*}[ht!]
	\begin{center}
		\includegraphics[width=0.98\textwidth]{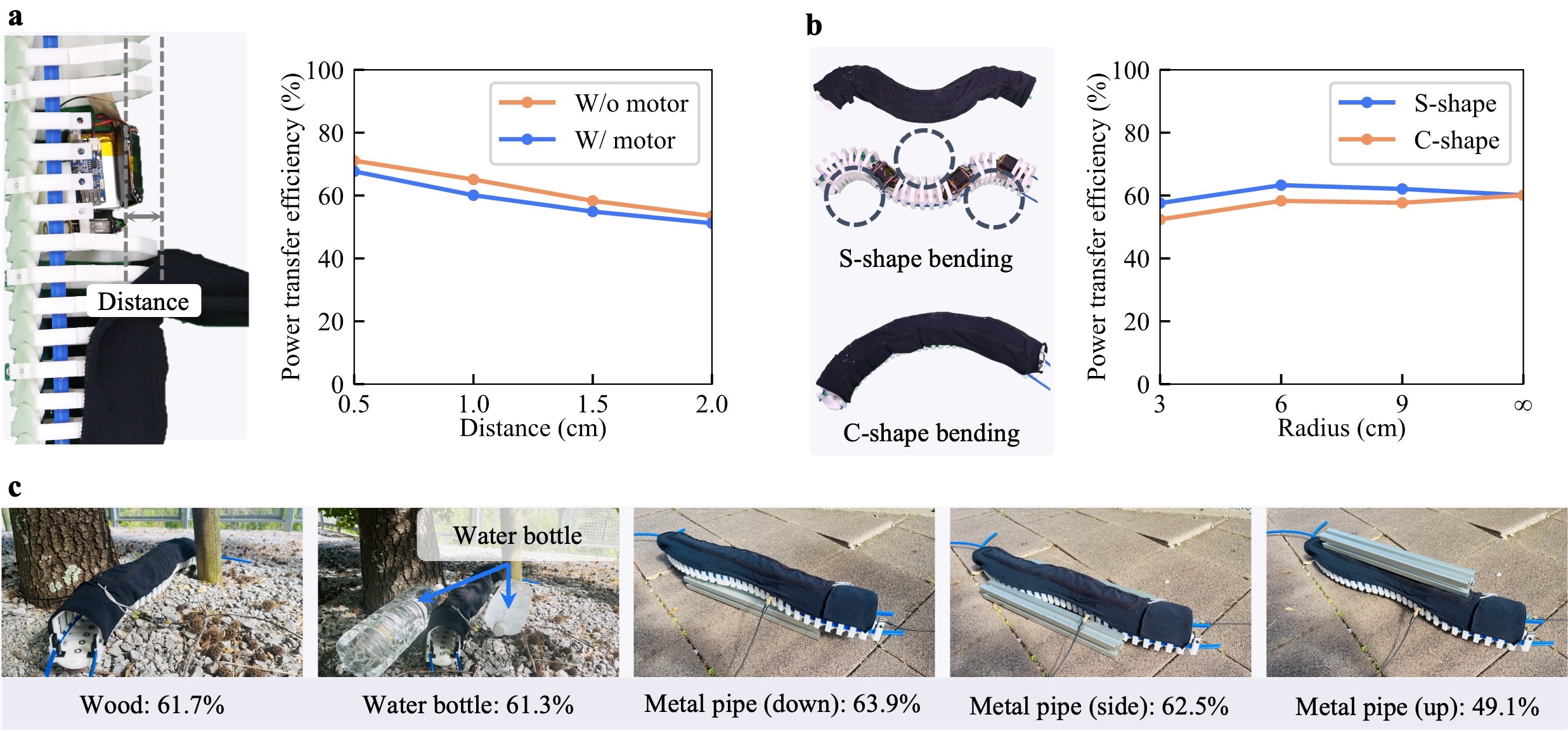}
	\end{center}
	\caption{\textbf{Wireless charging capability of the robot skin.} AC-to-AC power transfer efficiency is measured for (a) varying distances between TX/RX coils, (b) snake bending postures, and (c) performance in different surrounding environments.}
	\label{fig:eva_eff}
\end{figure*}

First, we discuss how to approximate the serpentine movement by the variable-length arc-shaped joint model, as explained in \S~\ref{sec:model}.
The serpentine movement basically follows a serpentine curve, whose curvature $\kappa$ varies sinusoidally.
The $\kappa$ can be expressed as follows:
\begin{align}
\kappa(t, s)=\frac{\pi \alpha_0}{2l_{\rm serp}} \sin \left(\omega t-\frac{\pi s}{2 l_{\rm serp}}\right)
\label{eq:kappa}
\end{align}
where $t$ is the time, $\omega$ is the angular frequency, $s$ is the distance along the curve, $l_{\rm serp}$ is the length of one-quarter of the curve, and $\alpha(t,s)$ is the winding angle along the curve. Note that $\alpha_0$ represents $\alpha (0,0)$.
Since the serpenoid curve can be approximated by serial segments of circular arcs~\cite{yamada2010approximations}, we approximate the serpenoid curve by our variable-length arc-shaped joint model as follows:
First, the serpenoid curve is divided into $N$ arc segments with $L_i$ length of $i$-th segment.
Then, the arc angle $\theta_i$ can be calculated as follows with $l = L_{\rm all}/4$ and \autoref{eq:kappa}:
\begin{align}
&\theta_i = \int_{\sum_{j=1}^{i-1} L_j}^{\sum_{j=1}^{i} L_j} \kappa(t, s) \, ds \quad(i=1,2,...,N) \notag\\
&= 2 \alpha_0 \sin\left(\frac{\pi}{L_{\rm all}}L_i\right) 
\sin\left[ \omega t - \frac{2\pi}{L_{\rm all}} \left(\sum_{j=1}^{i-1}L_j + \cfrac{L_i}{2}\right) \right] \label{eq:serpentine_locomotion}
\end{align}
where $\sum_{j=1}^{0} L_j \coloneqq 0$.
Lastly, based on \autoref{eq:Ltheta_to_d}, we calculate $d$ for $\theta$ and $L$ from the arc segment $\#1$ to fit the $N$ arc segments into the target serpenoid curve.
Note that $d$ for the more rearward units draws a complex composite sine wave due to the effect of the summing terms.
$N$ arc segmentation of the serpenoid curve is flexible, thereby allowing the proposed snake robot to perform serpentine locomotion with joint-position-free various segmentation.

\subsection{Joint-position-based Obstacle-aided Locomotion}
\label{sec:Joint-position-based Obstacle-aided Locomotion}

Obstacle-aided locomotion has been widely used as the promising locomotion strategy of rigid snake robots in confined, narrow terrains~\cite{transeth2008snake,takanashi2022obstacle,kano2017tegotae}. 
However, despite their high compliance, underactuated snake robots may not be compatible with obstacle-aided locomotion. While compliance generally enables adaptation to obstacles~\cite{wang2023mechanical}, underactuated mechanisms restrict the DoF of robot postures in confined spaces due to the limited number of motors.
In contrast, our underactuated snake robot based on the joint reconfigurable mechanism can address this challenge with a simple control algorithm.
To enable the obstacle-aided locomotion with repositionable-joint units, our strategy is to fix the shape of certain segments, and then, shift this fixed shape from the front to the rear along the trunk~(see \autoref{fig:control}b). 
The fixed shape encountering an obstacle enables the robot to propel forward through the reaction force from the obstacle.
To maintain the shape of segments across a given section, the angular velocities of the motors on the left and right sides of each joint unit in that section must be equal. For example, the angular velocity of all motors on the left side should be $\dot{d}_{L}$, while the angular velocity of all motors on the right side should be $\dot{d}_{R}$. When holding the shape from the $j$-th to the $k$-th segment ($j < k$), the angular velocity $\dot{d}$ of each motor is given by the following expression.
\begin{align}
 \dot{d}_{2j-1} &= \dot{d}_{2j+1} = \cdots = \dot{d}_{2k-3} = \dot{d}_{2k-1}\notag\\
 \dot{d}_{2j} &= \dot{d}_{2j+2} = \cdots = \dot{d}_{2k-2} = \dot{d}_{2k}
\label{eq:obstacle-aided_locomotion}
\end{align}
\autoref{eq:obstacle-aided_locomotion} allows the proposed robot to smoothly change shape and propel itself forward in obstacle-rich environments. However, the robot cannot perform this locomotion continuously, as it depends on joint unit positions. When the units reach the rear end of the body, they must be reset to the front to resume propulsion. During this resetting process, the units can advance along the robot’s shape, allowing the position to be reset with minimal impact on the robot’s overall movement.

\section{Experiment}

\begin{figure*}[t!]
	\begin{center}
		\includegraphics[width=0.98\textwidth]{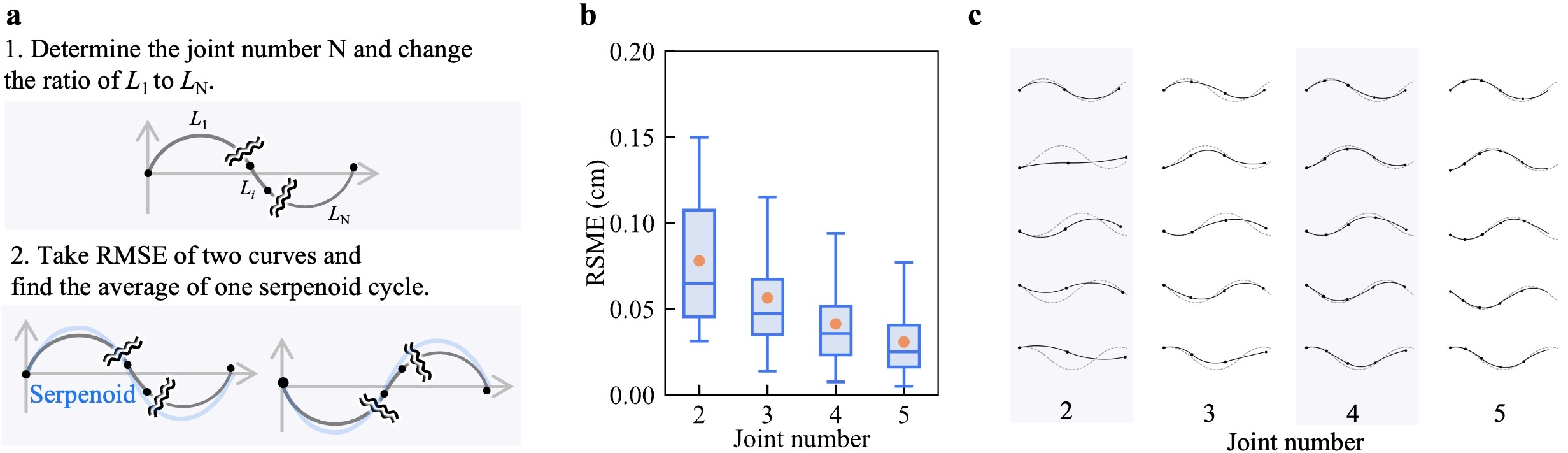}
	\end{center}
	\caption{\textbf{Simulation-based accuracy measurement of serpentine locomotion.} (a) Simulation protocol for measuring RMSE of a serpenoid curve using a variable-length arc-shaped joint model. (b) RMSE between the robot's and the serpenoid curves when varying the number of segments $N$. (c) Fitting result of a variable-length arc-shaped joint model with a different joint number for a target serpenoid curve.}
	\label{fig:simulation}
\end{figure*}
\begin{figure*}[t!]
	\begin{center}		
    \includegraphics[width=0.98\textwidth]{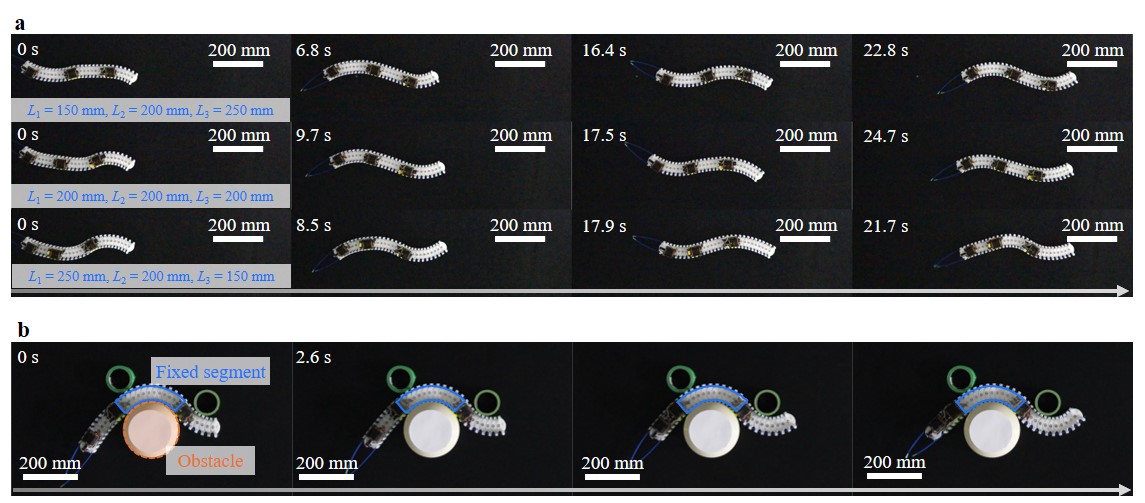}
	\end{center}
	\caption{\textbf{Demonstration of two types of locomotion control strategies.} (a) Time-lapse of joint-position-free serpentine locomotion. The robot's speed was almost the same in all three cases with different joint lengths. (b) Joint-position-based obstacle-aided locomotion. Based on the reaction force of a segment enclosed by a blue-colored line against three cylinder-shaped obstacles, the robot pushes forward.}
	\label{fig:locomotion}
\end{figure*}

\subsection{Wireless Charging Capability}

First, we investigate the wireless charging capability by measuring the AC-to-AC power transfer efficiency for the distance between the TX/RX coils, bending postures, and surrounding basic items.
The efficiency is calculated based on the measured S-parameter obtained from a vector network analyzer~(PicoVNA108).
\autoref{fig:eva_eff}a shows the efficiency for the distance between the TX/RX coils, ranging from \SI{0.5}{\cm} to \SI{2.0}{\cm} in steps of \SI{0.5}{\cm}.
The results show that the efficiency is almost the same with or without the joint units, indicating that the TX coil avoids the electromagnetic interference with the internal metallic motors.
Here, the distance of \SI{1}{\cm} is chosen to avoid collisions between the RX coil and the robot skin while keeping the high efficiency of approximately $60\%$.
Then, \autoref{fig:eva_eff}b illustrates the efficiency against the two types of bending posture: S-shape and C-shape with the three bending radii from \SI{3}{\cm} to \SI{9}{\cm} in steps of \SI{3}{\cm} and one straight line.
Owing to the relatively low $Q$-factor of the TX coil~($70$)~\cite{takahashi2022meander}, the efficiency constantly shows over $50\%$ for the different bending radius, enabling the joint units to stably receive \si{\W}-class power regardless of the robot bending. 
Lastly, \autoref{fig:eva_eff}c shows the efficiency against surrounding non-metallic~(\textit{i.e.}, wood, water bottle) or metallic items~(\textit{i.e.}, metal pipe).
The results show that the efficiency is almost the same except when the metal pipe above the TX coil causes electromagnetic interference, resulting in the efficiency drop of $13.4\%$.
The attachment of a ferromagnetic sheet above the robot skin could mitigate this electromagnetic interaction.
For further rigorous evaluation, the measurement in real-world testbeds such as disaster environments would be necessary.

\subsection{Simulation-based Accuracy of Serpentine Locomotion}
\label{sec:evaluation_RMSE}

Next, we evaluated the locomotion accuracy of the proposed robot. 
\autoref{fig:simulation} shows the simulation results of the fitting error for the variable-length arc-shaped joint model applied to a serpenoid curve. 
We set the total robot length to \SI{60}{\cm} varying the number of arc segments from $2$ to $5$. 
The length of each segment ranges from \numrange{2}{58}~\si{\cm}. 
As shown in \autoref{fig:simulation}a, the serpenoid curve ($l$: \SI{15}{\cm}, $\alpha_0$: \SI{0.7}{\radian}) was trimmed to match the robot’s length, and a series of points were plotted at equal intervals along both the robot and the curve. 
The total robot length matches one period of the serpenoid curve ($L_{\rm all} = 4l =$ \SI{60}{\cm}).
We calculated the root mean square error (RMSE) between these points to quantify the deviation. 
\autoref{fig:simulation}bc presents the RMSE and fitting pattern for the arc number. 
The result indicates that the fewer joints cause extremely long arc segments with large fitting errors for the serpenoid segment; the joint number over $3$ achieves approximate fitting accuracy.
Considering the trade-off between the fitting accuracy and the total motor number~(total robot weight), we chose $3$ joint units for prototype implementation. 
We would like to further investigate how variations in the robot's length influence the trade-off between the fitting accuracy and total robot weight,
both of which mainly influence the locomotion capability. 

\subsection{Locomotion Capability and Speed}
\label{sec:evaluation_locomotion}

Lastly, we demonstrated two types of locomotion as proof of concept.
\autoref{fig:locomotion}a shows a joint-position-free serpentine locomotion utilizing our prototype.
Based on the control law described in \S~\ref{sec:Joint-position-free Serpentine Locomotion}, the snake robot with different segment lengths moved along a serpenoid curve with $l = \SI{15}{\cm}$ and $\alpha_0 =\SI{0.70}{\radian}$, achieving a relatively fast velocity of \SI{2.25 \pm 0.05}{\cm/\s}.
The velocity is almost the same regardless of regardless of the unit's position. 
Additionally, \autoref{fig:locomotion}b shows a joint-position-based obstacle-aided locomotion. 
The prototype was placed in an obstacle-rich environment with three fixed cylinder-shaped obstacles. 
Following the control law described in \S~\ref{sec:Joint-position-based Obstacle-aided Locomotion}, the robot moved at a slow velocity of \SI{0.76}{\cm/\s}, while maintaining the segment shape in contact with the three obstacles. 
Note that all our current robot can do during obstacle-aided locomotion is move while maintaining a specified bending posture in a few obstacles.
To extend the current obstacle trajectories, one approach is to utilize sensors or cameras to recognize its bending posture relative to the surrounding obstacles and adjust accordingly to the real-world geometry.

\section{CONCLUSION}

This paper presents the joint-repositionable, inner-wireless snake robot capable of performing multi-joint, lightweight, low-powered basic locomotion.
The repositionable joint units enable underactuated multi-joint bending, in addition to the wireless charging-enables robot skin powering up the repositionable joint units continuously.
We studied the locomotion performance in the context of control mechanisms and wireless charging capability, demonstrating joint-repositionable locomotion with a wireless charging capability of \SI{7.6}{\W}.

Despite its promising features, the current simplified prototype can support only two types of planar slow locomotion, whereas the prior snake robots have demonstrated various fast locomotions for unstructured terrains~\cite{PEA-snake,AmphiBot,unified}. 
Our future research will focus on two key aspects: mechanism and control.
First, the development of multi-directional bending mechanism could allow the 3D locomotion to overcome unstructured terrains~\cite{mavinkurve2023geared}. 
Furthermore, the use of high-powered motors would be necessary for the fast locomotion of our robot.
Then, we will update the current model-based locomotion control to achieve adaptive locomotion in unstructured terrains.
The learning approach would be useful because various unique features of our robot including a flexible body, varying motor loads, and uneven surface contact force, make it hard to adjust the model-based locomotion for various terrains.
We strongly believe that our joint-repositionable inner-wireless snake robot could lead to even more practical robot explorers.

\addtolength{\textheight}{-3cm}   


\section*{ACKNOWLEDGMENT}

This work was supported by JST JPMJAX23K6, JPMJAX21K9, JPMJAP2401, JSPS 22K21343.
The authors would like to thank F. Kanada, T. Sato, and F. Kuroda for their help.

\bibliographystyle{IEEEtran}
\bibliography{reference}

\end{document}